# Computing L1 Straight-Line Fits to Data (Part 1)

Ian Barrodale[1]   *(ianbarrodale@gmail.com)*

# Introduction

In 1971 Frank D.K. Roberts and I developed an improved algorithm ([1],[2]), together with its Fortran implementation [3], for calculating L1 solutions to overdetermined systems of m linear equations in n unknown parameters[2]. In this technical report my plan is to first review some properties of L1 straight-line fits to data, and then to describe some algorithmic improvements to that 48-year old "BR" algorithm that are applicable to straight-line fitting. Hopefully, the description will be detailed enough to enable students (and others) to implement a BR for Line Fitting (BRLF) algorithm, in their preferred software development environment. Modern data analytics services can provide real-time analysis of fast data streams on a continuous basis, and code for a fast BRLF algorithm could be a useful addition for certain applications (e.g., [4]).

My initial remarks in this Part 1 report are primarily for those not familiar with the properties of L1 approximation, but the remainder of the report should also interest readers who are already acquainted with the inner workings of L1 algorithms. In Part 2 (in preparation) I shall deal with some additional algorithmic details (fast weighted median computations, expected locations of pivots, operation counts, duality, constrained solutions, etc.), and then review applications for which others have made good use of the original BR method and its Fortran code.

# 1 Background to L1 line fitting

**1.1 Prologue**  L1 refers here to a norm defined by *sums of absolute differences* between given data and an approximation to that data. Since this report concerns just the two-parameter case of fitting straight lines to data, the simplest example would be fitting a straight line to three distinct data points, in which case the best L1 approximation passes through the two end points. If the data set contains four distinct points, then matters may not be quite so simple. For example, there may be infinitely many best L1 straight-line approximations, as is the case with the data set {(1,0), (2,1), (3,1), (4,0)}. Since it is known that for any data set a best L1 straight line exists that passes through (*interpolates*) at least two of its points, with a four-point data set we need only choose between six lines at most. This is not a practical technique when fitting straight lines to larger data sets[3], since computational experience with [3] shows that usually relatively few pairs of data points need be examined. Even for million-point data sets used in evaluating modifications described in this report, solutions were obtained much faster than with [3] and typically involving only about a dozen pairs of candidate interpolation points.

---

[1] Adjunct Professor of Computer Science, University of Victoria, Victoria BC Canada.
[2] Google Scholar currently lists more than 1,250 citations to this algorithm and software ([2] and [3]).
[3] Nevertheless, those with access to a massively parallel computer might like to experiment with this approach.

### 1.2 The original BR algorithm

This algorithm is an adaptation of the Simplex method of linear programming (LP), and the enhancements described herein are designed to improve the performance of the BR method when computing straight-line fits to today's much larger data sets than those of five decades ago. The Simplex method was invented by George Dantzig in 1947, and although more recent algorithms execute faster on some types and sizes of LP problems, there is still good reason to continue using a version of the Simplex method for fitting L1 straight lines to real-world data sets.

The current unprecedented and exponential growth in number, size and type of data sets (including sensor data from industrial equipment, cars, aircraft, wearable and medical devices, smart meters, satellites, etc., plus data from social media sites, smart phones, browser logs, financial transactions, etc.) provides a challenging and diverse source of test bed material for developing alternative L1 data cleaning, interpretation and prediction methods for many aspects of our lives that are, and increasingly will be, affected by this data explosion.

### 1.3 Statement of the problem

The *weighted L1 straight-line approximation problem* is, for m given data values $(t_i, d_i)$, to minimize the sum of absolute residuals (SAR), where

$$SAR(a_1, a_2) = \sum w_i |d_i - (a_1 + a_2 t_i)|, \text{ for all } i = 1, 2, \ldots, m, \text{ and the } w_i \text{ are given positive weights.} \quad (1)$$

In practice the weights are often all set to 1, but it can be useful sometimes to assign larger weights to certain data. For example, when computing an L1 trend line to a sequence of data values $d_i$ recorded at each $t_i$ (times, typically), we might weight recent data more heavily than older data, more accurate data than less accurate data, or more relevant data to a task at hand than less relevant data (but which may still need to be considered). The choice of weights can have considerable influence on straight-line approximations; even the simple example in Section 1.1 involving just three points can have a different outcome when the middle weight is increased sufficiently. The Fortran code [3] for the original BR method has no built-in provision for user-supplied weights; rather, the user must pre-multiply each row of the initial Simplex tableau by its appropriate weight before a weighted L1 computation can proceed. Although this stratagem is very inefficient compared to the alternative (given in Section 2.8), it can be used to check for correctness when coding a faster built-in weighting scheme for use with BRLF.

### 1.4 Uniqueness of the solution

If a unique minimum of $SAR(a_1, a_2)$ is achieved for parameter values $a_1 = a_1^*$ and $a_2 = a_2^*$, then it is simple to prove (via the Simplex method, for example) that at least two of its m *residuals* $r_i = d_i - (a_1^* + a_2^* t_i)$ will be zero. If on the other hand a minimum value $SAR^* = SAR(a_1^*, a_2^*)$ is achieved for some other pair $(a_1, a_2)$, then by convexity it transpires that there will be a finite number of best L1 straight-line approximations having at least two zero residuals, together with infinitely many best L1 straight-line approximations with one or no zero residuals. For example, for the set {(1,0), (2,1), (3,1), (4,0)}, with all $w_i = 1$, any line intersecting both the line segment joining (1,0) and (2,1) and the line segment joining (3,1) and (4,0) is a best L1 line fit, as demonstrated next.

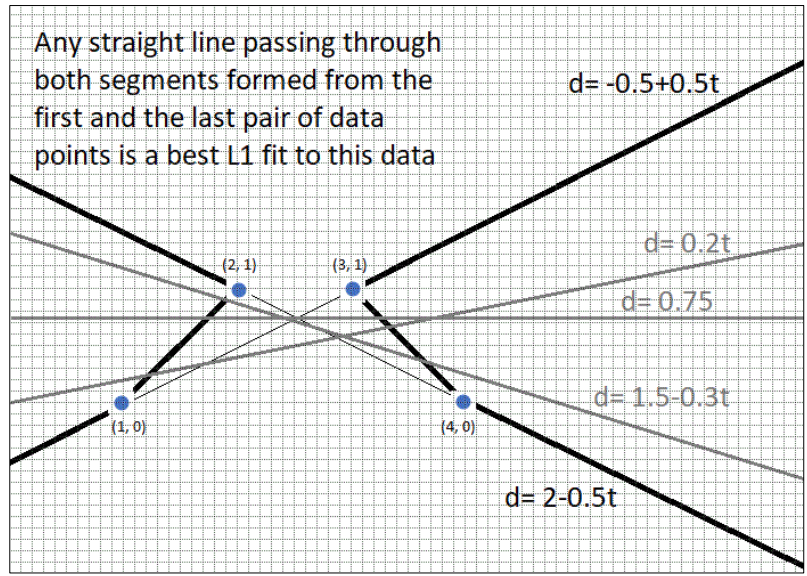

For this contrived data set there are four different best L1 straight-line approximations having two zero residuals (the two not shown are d = 0 and d = 2), and infinitely many with one or no zero residuals[4]. Furthermore, perturbing the d values above to produce the data set {(1,2), (2,3), (3,6), (4,4)}, it can be seen in the next figure that there would then be just three different best L1 straight-line approximations having two zero residuals, although still infinitely many with one or no zero residuals.

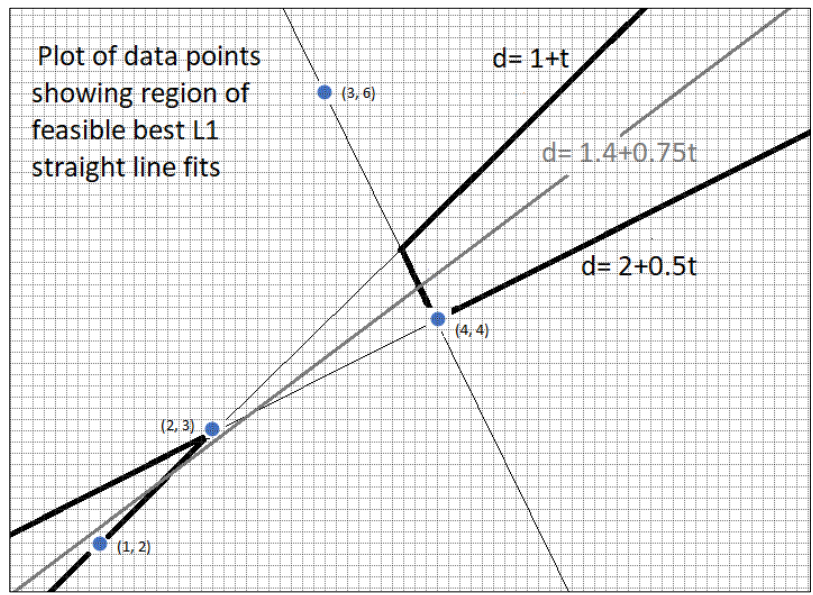

---

[4] In [8] it is shown that if Z denotes the number of zero residuals of a best L1 approximation (by *any* function, not just a straight line), P is the number of its positive residuals and N is the number of its negative residuals, then zero is a median of the m residuals if and only if $|P - N| \leq Z$, and for m odd, $|P - N| \leq Z-1$. For straight line L1 fits to real data sets it is usually the case that Z = 2, so the residuals are well balanced in sign (while often not so in magnitude).

Even for real (non-contrived) data sets, the issue of non-uniqueness does sometimes occur in L1 approximation[5]. Section 2.2 explains how to record these occurrences (with more definiteness than in the original BR Fortran implementation, which after convergence just warned whenever the computed solution was *probably* non-unique).

### 1.5 Choosing Simplex method pivots

*Medians* and *weighted medians* can be used to advantage in choosing pivots, although some confused claims by others have been made in the past about the role of weighted medians in the original BR method.

To clarify matters, setting all $w_i$ = 1 and $a_2$ = 0 in expression (1) above, the resulting one parameter problem is to minimize $F(a_1) = \sum |d_i - a_1|$ for all i = 1,2,…,m, and a solution $a_1^*$ to this L1 problem is a median of the numbers $d_1, d_2, …, d_m$. Now when $a_1$ is eliminated instead of $a_2$, a solution $a_2^*$ to this one parameter problem of minimizing $F(a_2) = \sum |d_i - a_2 t_i|$ for all i = 1,2,…,m is a solution to the weighted L1 problem of minimizing $\sum |t_i| |(d_i/t_i) - a_2|$. Thus, this solution $a_2^*$ is a weighted median of finite ratios $(d_i/t_i)$ with positive weights $|t_i|$.

The rule given for choosing a pivot in each Simplex iteration of the original BR method makes no reference to weighted medians, and (to eliminate confusion) the BR choice of pivot is *not always* equivalent to choosing a certain weighted median. Although a weighted median strategy can considerably reduce the computational effort involved in choosing pivots (when compared to the BR rule), it can also *increase* the number of iterations before convergence occurs, and more seriously, it can cause iterations to *cycle without ever converging* to a best L1 solution (as shown later by examples). So, if pivots are chosen using weighted medians in any BRSL implementation, it is *essential* to employ a fail-safe strategy combining an efficient weighted median algorithm with the guaranteed convergence provided by the original BR pivot selection method.

### 1.6 Using a trial solution

Another enhancement that can be effective is the option of beginning the Simplex iterations from a trial solution, rather than from a cold start that assumes no previous knowledge. For straight-line fitting one option is to use the best least-squares (L2) parameters and residuals, as follows. The (unique) least-squares values for parameters $a_1$ and $a_2$ are first computed from standard formulas[6], the corresponding L2 residuals are calculated, and these are fitted by an L1 straight line; the final parameters $a_1^*$ and $a_2^*$ are then obtained from the L2 fit (to the original data) and the L1 fit (to the L2 residuals) by adding together both intercept values and both slope values. Usually, but not always, this extra work is compensated for by a reduction in the subsequent number of iterations required to arrive at an L1 solution. BRSL users should also have the option of starting the iterations from already-available L2 parameters and residuals, or *any other* inputted trial parameters and corresponding residuals, always with the aim of reducing the overall computational effort. For small data sets, where convergence from a cold start with BR often occurs in five or fewer iterations, reductions in overall effort can be less impressive unless good trial parameters and corresponding residuals are already available (i.e., with little or no pre-computation). Of course, the original BR Fortran code [3] does not *have* to start cold, but it has no built-in option for a fast start. That must be arranged by user-supplied input of parameter and residual values obtained from a trial solution.

---

[5] Non-uniqueness of L1 straight-line fits seems to occur only rarely with larger data sets, and when the m values $t_i$ are equally spaced *and* m = 6+4k for any non-negative integer k, uniqueness is *guaranteed* (see Section 2.3).
[6] First calculate C1 = $\sum w_i t_i$; C2 = $\sum w_i d_i$; C3 = $\sum w_i t_i^2$; C4 = $\sum w_i t_i d_i$; C5 = $\sum w_i$; and let D = C1 C1 – C3 C5. Then the best L2 parameters are $a_2$ = (C1 C2 – C4 C5)/D and $a_1$ = (C1 C4 – C2 C3)/D. Note that if the RHS is changed only two of these L2 summations (C2 and C4) must be recalculated, provided that the $w_i$ and $t_i$ values remain unchanged.

1.7 Epilogue  This concludes the background summary, with insights into some of the modifications to the BR method that can lead to a more efficient algorithm, BRSL, for L1 fitting of data by straight lines. These enhancements can result in more versatile and faster software (particularly for large data sets and depending on the programming language involved) than Algorithm 478 in the Communications of the ACM, which is the Fortran subroutine [3]. Some speedup would be inevitable, because Algorithm 478 solves the general L1 approximation problem of fitting an n-parameter function to m given data points, where m ≥ n. It has no special provisions applicable just for straight-line fitting.

The next section contains additional remarks about L1 approximations, some computational results obtained to date, plus some nitty-gritty algorithmic and coding details.

# 2 Diving deeper into L1 line fitting

2.1 The original BR algorithm  The details of the BR algorithm have been in print since the early 1970s, so here is just a brief illustration (see [2] for more details) of that original algorithm calculating an L1 straight-line fit to the data set {(1,1), (2,1), (3,2), (4,3), (5,2)} with all the weights set to 1. Expressing problem (1) as an LP problem using the standard form of the Simplex method, the full initial Simplex tableau for this five-point example is displayed next, followed by its two-iteration solution using condensed tableaux.

| Costs → | | 0 | 0 | 0 | 0 | 1 | 1 | 1 | 1 | 1 | 1 | 1 | 1 | 1 | 1 | |
|---|---|---|---|---|---|---|---|---|---|---|---|---|---|---|---|---|
| ↓ | Basis | $b_1$ | $b_2$ | $c_1$ | $c_2$ | $u_1$ | $u_2$ | $u_3$ | $u_4$ | $u_5$ | $v_1$ | $v_2$ | $v_3$ | $v_4$ | $v_5$ | R |
| 1 | $u_1$ | 1 | 1 | -1 | -1 | 1 | | | | | -1 | | | | | 1 |
| 1 | $u_2$ | 1 | 2** | -1 | -2 | | 1 | | | | | -1 | | | | 1 |
| 1 | $u_3$ | 1 | 3*** | -1 | -3 | | | 1 | | | | | -1 | | | 2 |
| 1 | $u_4$ | 1 | 4 | -1 | -4 | | | | 1 | | | | | -1 | | 3 |
| 1 | $u_5$ | 1 | 5* | -1 | -5 | | | | | 1 | | | | | -1 | 2 |
| Marginal Costs→ | | 5 | 15 | -5 | -15 | 0 | 0 | 0 | 0 | 0 | -2 | -2 | -2 | -2 | -2 | 9 |

To use the Simplex method, nonnegative variables $u_i$, $v_i$, $b_j$, and $c_j$ have been introduced, putting $a_j = b_j - c_j$ and $d_i - (a_1 + a_2 t_i) = u_i - v_i$, and then minimizing expression (1) on page 2 when all $w_i = 1$ can be replaced by the following LP problem:

  minimize $\sum(u_i + v_i)$ subject to $d_i = (b_1-c_1) + (b_2-c_2)t_i + u_i - v_i$  for all i = 1,2,…,m.                    (2)

The columns of the Simplex tableau above have obvious names as vectors, except that **R** contains the $d_i$ values. Rather than the term Reduced Costs, the term Marginal Costs (MC) is used for the bottom row, as in [5] where it was also observed that an initial LP basic feasible solution to expression (2) is immediately available[7] and that most of the columns shown in the tableau above need not be stored explicitly. When required, hidden vectors can replace exhibited vectors with just a sign change, and the marginal costs of each pair $u_i$ and $v_i$ must sum to -2. The choice of vectors to enter the basis in the first two iterations is restricted to one of $b_1$ or $c_1$ and one of $b_2$ or $c_2$, and thereafter one of each pair[8] is kept in the basis. In each iteration thereafter, a $u_i$ or $v_i$ with the largest MC enters the basis. Most importantly, in every BR iteration (from first to last) a vector to leave the basis is chosen by using a pivot that guarantees the maximum possible decrease in the objective function $\sum (u_i + v_i)$. A solution for the data set {(1,1), (2,1), (3,2), (4,3), (5,2)} is obtained in two such iterations as exhibited in the next figure; the bottom entry below **R** is the current value of SAR for each iteration, and its minimum value SAR* = 2 is shown in that cell of the final tableau:

| Basis | $b_1$ | $b_2$ | R |
|---|---|---|---|
| $u_1$ | 1 | 1 | 1 |
| $u_2$ | 1 | 2** | 1 |
| $u_3$ | 1 | 3*** | 2 |
| $u_4$ | 1 | 4 | 3 |
| $u_5$ | 1 | 5* | 2 |
| MC-> | 5 | 15 | 9 |

| Basis | $b_1$ | $u_3$ | R |
|---|---|---|---|
| $u_1$ | 2/3* | -1/3 | 1/3 |
| $v_2$ | -1/3 | 2/3 | 1/3 |
| $b_2$ | 1/3 | 1/3 | 2/3 |
| $u_4$ | -1/3 | -4/3 | 1/3 |
| $v_5$ | 2/3 | 5/3 | 4/3 |
| MC-> | 2/3 | -1/3 | 7/3 |

| Basis | $u_1$ | $u_3$ | R |
|---|---|---|---|
| $b_1$ | 3/2 | -1/2 | 1/2 |
| $v_2$ | 1/2 | 1/2 | 1/2 |
| $b_2$ | -1/2 | 1/2 | 1/2 |
| $u_4$ | 1/2 | -3/2 | 1/2 |
| $v_5$ | -1 | 2 | 1 |
| MC-> | -1 | 0 | 2 |

The vector $b_2$ with the largest marginal cost is initially selected to enter the basis, the first and second pivot choices (indicated by * and **) are bypassed, and 3 is chosen as the optimal pivot, which reduces $\sum(u_i + v_i)$ from 9 to 7/3. In the second tableau the first pivot in $b_1$ is optimal, and after two iterations the solution d = 0.5 + 0.5t is obtained with $\sum(u_i + v_i)$ = 2. Convergence is assured because all 2m vectors $u_j$ and $v_j$ satisfy -2 ≤ MC ≤ 0, while $b_1$, $b_2$, $c_1$, and $c_2$ have MC = 0. Bypassing pivots typically reduces iteration counts substantially, and when combined with other stratagems in this report, a BRSL user should be able to fit L1 straight lines to large data sets without experiencing the frustration and inconvenience of sluggish execution times.

2.2 The Simplex method and non-uniqueness  Having already provided some examples of non-unique L1 solutions, let us recall how the Simplex method indicates non-uniqueness. It can only occur when in the final tableau a vector that is not in the final basis has a marginal cost of zero *and* the ratio test to choose a pivot from that vector is positive. A pivot chosen from a zero ratio will not change the current values for $a_1$ or $a_2$. In the third (final) condensed tableau in Section 2.1, notice that $u_3$ can replace $v_5$ in the basis by choosing 2 as the pivot, and this additional iteration produces an alternate optimal L1 solution of z = 0.75 + 0.25t.

---

[7] Albeit with $b_1 = c_1 = b_2 = c_2 = 0$. This LP basic feasible solution is not a basic L1 solution (unless all $d_i$ = 0). For straight-line fitting we define *a basic L1 solution as one that interpolates at least two data points.*

[8] An exception can occur when the algorithm converges after just one iteration because an L1 straight-line fit can be attained through use of a single parameter, as happens when the d values are all equal, or if they all lie on a straight line through the origin. A one-iteration solution of d = 10.47t occurs with the erratic nine-point data set {(4, 291.3), (5,-107.1), (6,-104.6), (7,97.8), (8,-100), (9,302.8), (10,104.7}, (11, 307), (12,-90.9)}; this set also has two basic L1 solutions, each interpolating two data points, and all three of these L1 solutions interpolate (10,104.7).

## 2.3 Chances of unique vs non-unique solutions

What is known about the likelihood of uniqueness or non-uniqueness of L1 straight line fits *before processing begins* on a data set? In general, it is not possible to make such predictions in advance, but some interesting results that apply *just to equally spaced data* are as follows. It is shown in [7] that provided the t values are equidistant, (a) uniqueness is guaranteed whenever m is even but not a multiple of 4, (b) the chance of a non-unique solution when m = 8 is at least 3%, (c) for any m the lower bound for the chance of a non-unique L1 line is approximately $(2.2)/m^2$ (which is 0.55% when m = 20 and 0.022% when m = 100), (d) when m is odd there are at most two basic solutions, and (e) when m is a multiple of 4 the solution is either unique or there are more than two basic L1 solutions. Evidently, unless m = 6+4k, non-uniqueness is not uncommon for *small* equally spaced data sets.

## 2.4 The CPI data set

Most of the numerical examples from here onwards make use of the 21 equally spaced data below representing the annual Consumer Price Index (CPI) for Canada, from 1995 (t = 1) to 2015 (t = 21), where d = 100 in 2002[9]. (UK CPI data was used similarly in [7].)

| t | d | t | d | t | d |
|---|---|---|---|---|---|
| 1 | 87.6 | 8 | 100 | 15 | 114.4 |
| 2 | 88.9 | 9 | 102.8 | 16 | 116.5 |
| 3 | 90.4 | 10 | 104.7 | 17 | 119.9 |
| 4 | 91.3 | 11 | 107 | 18 | 121.7 |
| 5 | 92.9 | 12 | 109.1 | 19 | 122.8 |
| 6 | 95.4 | 13 | 111.5 | 20 | 125.2 |
| 7 | 97.8 | 14 | 114.1 | 21 | 126.6 |

Starting at t = 1 with a sequence of length 4 (data set L4S1), then at t = 2 with a sequence of length 4 (L4S2), through to t = 18 with a sequence of length 4 (L4S18), and then at t = 1 with a sequence of length 5 (L5S1), through to t = 17 with a sequence of length 5 (L5S17), and so on until the last data set L21S1 was formed, a total of 171 data sets were assembled. L1 straight-line fits were then computed for all 171 data sets, and the incidence of uniqueness and non-uniqueness, and each optimal value SAR*, are contained in the next two figures.

---

[9] The CPI for Canada is a composite price index that compares prices for consumer products in various price observation periods to prices in the index base period; when last adjusted the base was set to 100 in 2002.

| Sequence Start → | 1 | 2 | 3 | 4 | 5 | 6 | 7 | 8 | 9 | 10 | 11 | 12 | 13 | 14 | 15 | 16 | 17 | 18 |
|---|---|---|---|---|---|---|---|---|---|---|---|---|---|---|---|---|---|---|
| Length ↓ | | | | | | | | | | | | | | | | | | |
| 4 | N | U | N | N | N | N | U | N | U | U | N | N | U | N | U | N | U | U |
| 5 | U | N | U | N | N | U | U | N | N | N | N | N | N | N | N | N | U | |
| 6 | U | U | U | U | U | U | U | U | U | U | U | U | U | U | U | U | | |
| 7 | U | U | U | U | U | N | U | N | U | U | U | U | U | U | U | | | |
| 8 | N | N | N | N | U | U | U | U | N | U | U | N | U | U | | | | |
| 9 | U | U | U | U | U | U | U | U | N | U | U | N | U | | | | | |
| 10 | U | U | U | U | U | U | U | U | U | U | U | U | | | | | | |
| 11 | U | U | U | N | U | N | U | U | U | U | N | | | | | | | |
| 12 | U | U | N | U | U | N | U | U | U | U | | | | | | | | |
| 13 | U | U | U | U | U | U | U | U | U | | | | | | | | | |
| 14 | U | U | U | U | U | U | U | U | | | | | | | | | | |
| 15 | U | U | U | U | N | U | U | | | | | | | | | | | |
| 16 | U | U | U | U | U | U | | | | | | | | | | | | |
| 17 | U | U | U | U | U | | | | | | | | | | | | | |
| 18 | U | U | U | U | | | | | | | | | | | | | | |
| 19 | U | U | U | | | | | | | | | | | | | | | |
| 20 | U | U | | | | | | | | | | | | | | | | |
| 21 | U | | | | | | | | | | | | | | | | | |

Table header: Uniqueness (U) or Non-uniqueness (N) of L1 line fits to 21-point CPI data

For each cell, if sequence length is:
- Multiple of 4 → 1 or >2 basic L1 solutions
- Odd number → 1 or 2 basic L1 solutions
- Even number ≠ 4*k → Unique L1 solution

An example of the multiple solution possibilities in the framed summary is on the next page.

| Sequence Start → | 1 | 2 | 3 | 4 | 5 | 6 | 7 | 8 | 9 | 10 | 11 | 12 | 13 | 14 | 15 | 16 | 17 | 18 |
|---|---|---|---|---|---|---|---|---|---|---|---|---|---|---|---|---|---|---|
| Length ↓ | | | | | | | | | | | | | | | | | | |
| 4 | 0.40 | 0.43 | 1.60 | 0.80 | 0.30 | 0.40 | 0.50 | 0.50 | 0.20 | 0.17 | 0.50 | 2.10 | 1.37 | 3.10 | 0.97 | 2.30 | 0.67 | 0.77 |
| 5 | 0.45 | 1.60 | 1.95 | 0.80 | 0.40 | 0.55 | 0.50 | 0.70 | 0.50 | 0.50 | 2.10 | 2.10 | 3.10 | 3.10 | 2.30 | 2.30 | 0.85 | |
| 6 | 1.63 | 2.50 | 2.10 | 1.20 | 0.73 | 0.58 | 0.70 | 0.70 | 0.85 | 2.13 | 2.35 | 3.34 | 3.10 | 3.74 | 2.30 | 2.58 | | |
| 7 | 2.95 | 3.40 | 2.75 | 1.28 | 0.75 | 0.80 | 0.75 | 1.00 | 2.30 | 2.90 | 3.35 | 3.40 | 3.88 | 3.85 | 3.00 | | | |
| 8 | 4.30 | 4.70 | 3.10 | 1.30 | 0.98 | 0.87 | 1.00 | 2.67 | 3.30 | 3.48 | 3.40 | 4.40 | 3.96 | 4.17 | | | | |
| 9 | 5.60 | 5.05 | 3.10 | 1.50 | 1.10 | 1.08 | 2.73 | 4.00 | 3.60 | 3.60 | 4.40 | 4.40 | 4.38 | | | | | |
| 10 | 6.60 | 5.23 | 3.23 | 1.55 | 1.26 | 2.85 | 4.50 | 4.12 | 3.60 | 4.60 | 4.49 | 4.89 | | | | | | |
| 11 | 7.47 | 5.28 | 3.27 | 1.80 | 3.10 | 4.70 | 4.59 | 4.30 | 4.60 | 4.83 | 5.10 | | | | | | | |
| 12 | 7.79 | 5.40 | 3.60 | 3.53 | 5.00 | 5.20 | 4.96 | 5.30 | 4.99 | 5.46 | | | | | | | | |
| 13 | 8.20 | 5.83 | 5.23 | 5.36 | 5.58 | 5.57 | 6.08 | 5.81 | 5.61 | | | | | | | | | |
| 14 | 8.86 | 7.33 | 6.83 | 5.80 | 6.09 | 6.88 | 6.74 | 6.45 | | | | | | | | | | |
| 15 | 10 | 8.88 | 7.10 | 6.20 | 7.60 | 7.63 | 7.37 | | | | | | | | | | | |
| 16 | 10.7 | 8.99 | 7.50 | 7.60 | 8.44 | 8.50 | | | | | | | | | | | | |
| 17 | 11.3 | 9.36 | 8.87 | 8.46 | 9.70 | | | | | | | | | | | | | |
| 18 | 11.3 | 10.5 | 9.71 | 10 | | | | | | | | | | | | | | |
| 19 | 12.3 | 11.2 | 10.9 | | | | | | | | | | | | | | | |
| 20 | 12.5 | 11.8 | | | | | | | | | | | | | | | | |
| 21 | 13.1 | | | | | | | | | | | | | | | | | |

Table header: Sum (to 2D) of Absolute Residuals (SAR* values) for L1 line fits to 21-point CPI data

As noted later in Section 2.9, there are 12 SAR* values here (e.g., for L4S4) where SAR* does not increase when the next point in the sequence is added to the data (L5S4)

The figure below provides an illustration of the framed summary in the top figure on the previous page, for m = 4, 5, and 6, using the equally spaced data sets L4S4, L5S4, and L6S4. The labels identifying vectors in these and all future tableaux are as follows: **b₁** has the label 1, **b₂** has label 2, **u₁** has label 3, **u₂** has label 4,…, **u_m** has label (m+2), and the **c_j** and **v_i** vectors have the corresponding negative integers as their labels.

| Multiple of 4 | | ① | | | | ② | | | | ③ | | | |
|---|---|---|---|---|---|---|---|---|---|---|---|---|---|
| t | d | Final Results | | | | Alternate Solution | | | | Alternate Solution | | | |
| 4 | 91.3 | -0.5 | -0.5 | 2.05 | 2 | 1 | 1 | 2.4 | 2 | -0.33333 | 0.33333 | 2.16667 | 2 |
| 5 | 92.9 | 3 | 2 | 83.1 | 1 | -6 | -7 | 81 | 1 | 2.33333 | -1.33333 | 82.6333 | 1 |
| 6 | 95.4 | 0.5 | -0.5 | 0.45 | -4 | -1 | -2 | 0.1 | -4 | 0.66667 | 0.33333 | 0.56667 | -4 |
| 7 | 97.8 | 0.5 | 1.5 | 0.35 | 6 | 2 | 3 | 0.7 | 3 | 0.33333 | 0.66667 | 0.23333 | -5 |
| | | 0 | 0 | 0.8 | | 0 | 0 | 0.8 | | 0 | 0 | 0.8 | |
| | | 3 | -5 | | | 6 | -5 | | | 3 | 6 | | |

| Odd number | | ④ | | | | ⑤ | | | | | | | |
|---|---|---|---|---|---|---|---|---|---|---|---|---|---|
| t | d | Final Results | | | | Alternate Solution | | | | | | | |
| 4 | 91.3 | -0.5 | 0.5 | 2.3 | 2 | -1 | -1 | 2.2 | 2 | | | | |
| 5 | 92.9 | 3 | -4 | 81.6 | 1 | 7 | 8 | 82.4 | 1 | | | | |
| 6 | 95.4 | -1 | 2 | 0.5 | 3 | -3 | -4 | 0.1 | 3 | | | | |
| 7 | 97.8 | 0.5 | 0.5 | 0.1 | 6 | 2 | 2 | 0.2 | -5 | | | | |
| 8 | 100 | 0.5 | -1.5 | 0.2 | -4 | 2 | 3 | 0.5 | -4 | | | | |
| | | -1 | 0 | 0.8 | | -1 | -2 | 0.8 | | | | | |
| | | -7 | -5 | | | -7 | 6 | | | | | | |

| Even but not 4k | | ⑥ | | | | | | | | | | | |
|---|---|---|---|---|---|---|---|---|---|---|---|---|---|
| t | d | Final Results | | | | | | | | | | | |
| 4 | 91.3 | -1 | 1 | 2.4 | 2 | | m is a multiple of 4 | ⇒ | 1 or >2 basic L1 solutions | | | | |
| 5 | 92.9 | 7 | -6 | 81 | 1 | | | | | | | | |
| 6 | 95.4 | 2 | -1 | 0.1 | -4 | | m is an odd number | ⇒ | 1 or 2 basic L1 solutions | | | | |
| 7 | 97.8 | -3 | 2 | 0.7 | 3 | | | | | | | | |
| 8 | 100 | -1 | 2 | 0.2 | -7 | | m is even but not 4k | ⇒ | Unique L1 solution | | | | |
| 9 | 102.8 | 2 | -3 | 0.2 | 8 | | | | | | | | |
| | | -1 | -1 | 1.2 | | | | | | | | | |
| | | 5 | 6 | | | | | | | | | | |

Note in this figure that both columns in Table 1 have zero marginal costs, and Tables 2 and 3 result by pivoting appropriately. Table 4 has one column with MC = 0, and Table 5 is the result, while Table 6 has no column with MC = 0.

## 2.5 Is it positive or negative?

It is appropriate here to introduce a nitty-gritty coding detail. When implementing BR as Algorithm 478 [3], a positive number TOLER (typically set to $10^{-11}$ for double precision[10] computations with well-scaled[11] data sets) was used for *decision making* in the code: any number z for which |z| ≤ TOLER could be considered to have the value zero. A number z for which z > TOLER was positive, and a number z for which z < (-TOLER) was negative.

---

[10] At that time double precision implied 64 bits, giving about 16 decimal digits of precision. Programming languages now allow code to be written using multiple precisions (single, double, quad, and/or some other precision) without syntax changes, so a more sophisticated strategy than using a single-valued TOLER could be employed today.

[11] When warranted (e.g., for data sets with some extremely large and disparate deviations from zero), the $t_i$-values and $d_i$-values could first be normalized by dividing by max|$t_i$| and max|$d_i$|, respectively. Recovering appropriate unscaled values from the final tableau involves multiplication of all m elements of **R** by max|$d_i$|, followed by division of the resulting $a_2$ value by max|$t_i$| to yield an unscaled optimal value $a_2^*$.

Consequently, a candidate to enter the basis should be chosen only from vectors whose marginal costs are non-negative (≥ (-TOLER)), and pivot candidates from the vector to enter the basis should be positive (> TOLER). If, due to limitations of floating-point arithmetic, no positive pivot can be found before convergence occurs (an extremely rare occurrence in practice), then the algorithm should be terminated prematurely. Though much in the world of computer hardware and software has been transmuted in the past half century, sometimes the need to distinguish carefully between calculated positive, negative and (essentially) zero values is inescapable.

An illustration of this need is provided below by the CPI data set L6S7. Changing signs in rows with negative entries in **R** is necessary to maintain LP feasibility as the iterations proceed. Convergence occurs as soon as the marginal costs of all 2m vectors associated with the nonnegative variables $u_i$, and $v_i$ lie in the range [-2, 0]. If an entry in **R** is judged to have the value zero (a *degeneracy* in LP terminology), its row can be multiplied by -1 without changing the current value of SAR. This action is warranted (as is the case with data set L6S7 shown below) if subtracting twice that row (before multiplication by -1) from the current marginal costs results in values in the range [-2, 0], thereby eliminating at least one additional iteration (a worthwhile saving with any very large data set).

| Initially (pivots in boxes) | | | | After one iteration | | | | After two iterations | | | | Final Results (after 3 iterations) | | | |
|---|---|---|---|---|---|---|---|---|---|---|---|---|---|---|---|
| 1 | 7 | 97.8 | 3 | 0.1 | 0.1 | 10.47 | 2 | -0.33333 | 0.333333 | 2.3 | 2 | -0.25 | -0.25 | 2.3 | 2 |
| 1 | 8 | 100 | 4 | 0.2 | -0.8 | 16.24 | 4 | 3.333333 | -2.33333 | 81.7 | 1 | 2.75 | 1.75 | 81.7 | 1 |
| 1 | 9 | 102.8 | 5 | 0.1 | -0.9 | 8.57 | 5 | -0.33333 | -0.66667 | 0.4 | 5 | -0.5 | 0.5 | 0.4 | 5 |
| 1 | 10 | 104.7 | 6 | 0.3 | -0.7 | 24.51 | 3 | 0.666667 | 0.333333 | 0.1 | -4 | 0.75 | -0.25 | 0.1 | -4 |
| 1 | 11 | 107 | 7 | 0.1 | 1.1 | 8.17 | -7 | -0.33333 | 1.333333 | -8.9E-15 | -7 | -0.25 | 0.75 | -6.7E-15 | 6 |
| 1 | 12 | 109.1 | 8 | 0.2 | 1.2 | 16.54 | -8 | -0.66667 | 1.666667 | 0.2 | -8 | -0.25 | -1.25 | 0.2 | -8 |
| 6 | 57 | 621.4 | | 0.9 | -1.1 | 74.03 | | -1.66667 | 1.666667 | 0.7 | | -1.25 | -1.25 | 0.7 | |
| 1 | 2 | | | 1 | 6 | | | 3 | 6 | | | 3 | -7 | | |

| | | | | | | Swap 5th row & test marginal costs | | | |
|---|---|---|---|---|---|---|---|---|---|
| | | | | | | Final Results (after 2 iterations) | | | |
| For organizational purposes, as | | | | | | -0.33333 | 0.333333 | 2.3 | 2 |
| the intercept and slope vectors | | | | | | 3.333333 | -2.33333 | 81.7 | 1 |
| enter the basis, they are stored | | | | | | -0.33333 | -0.66667 | 0.4 | 5 |
| in the top two rows. Thereafter | | | | | | 0.666667 | 0.333333 | 0.1 | -4 |
| they never leave the basis. | | | | | | 0.333333 | -1.33333 | 8.90E-15 | 7 |
| | | | | | | -0.66667 | 1.666667 | 0.2 | -8 |
| | | | | | | -1 | -1 | 0.7 | |
| | | | | | | 3 | 6 | | |

## 2.6 Incorporating L2 solutions into L1 calculations

The original BR Algorithm 478 [3] converges in either 2 or 3 iterations for almost 70% of the 171 CPI data sets, and either 4 or 5 iterations are required in the remainder of these test cases. An example of how the use of the least-squares (L2) solution can reduce iteration counts is provided by L9S4, which is one of the data sets needing 5 iterations to converge (to the L1 solution d = 81.7 + 2.3t with SAR* = 1.5). The L2 solution to L9S4 has an intercept of 81.831111 and a slope of 2.285, for which the L2

Computing L1 Straight-Line Fits to Data (Part 1)                                       December 2019residuals at t = 4, 5, 6,..., 12 are 0.328889, -0.356111, -0.141111,...,-0.151111. The L1 solution to this data set of L2 residuals converges in just 2 iterations with an intercept of -0.131111 and a slope of 0.015, which when added to the L2 intercept and slope provides the same L1 solution as that obtained directly. In this case the extra work involved in first computing the L2 solution and its residuals is compensated for by the reduction in L1 iterations from five to two. These savings in computational effort become more impressive with larger data sets, particularly when L2 solutions are being computed beforehand anyway (e.g., when both norms are used to compare solutions, say, because the data may contain *outliers* ([4]).

## 2.7 Pivots determined as weighted medians

Data set L10S12 provides another example where using L2 information at the start of the L1 calculations reduces the number of Simplex iterations required (three are sufficient here), but it also shows that BR pivots (five are needed from a cold start) can differ from those determined by the weighted median (WM) rule[12] (seven iterations are needed from a cold start). Here are the details; the unique L1 solution shown on either of the first two lines below the header can be obtained in fewer iterations by adding the slope and intercept values shown on the last two lines:

| Data set L10S12 | Slope | Intercept | SAR* | No. Iters |
|---|---|---|---|---|
| L1 fit (WM pivots) | 1.887500 | 86.962500 | 4.887500 | 7 |
| L1 fit (BR pivots) | 1.887500 | 86.962500 | 4.887500 | 5 |
| L2 fit (see footnote 6) | 1.952727 | 85.960000 | -- | -- |
| L1 fit to L2 residuals | -0.065227 | 1.002500 | 4.887500 | 3 |

Reference was made earlier in Section 1.5 to drawbacks that can occur when pivots are picked using a weighted median strategy alone. Data set L17S3 provides an example where the SAR value *increases* in the fourth iteration, although convergence does occur in the fifth iteration. BR takes one less iteration to reach the same SAR* value of 8.870.

| Data set L17S3 | BR pivots | | WM pivots | | |
|---|---|---|---|---|---|
| Iter no. | Interp pts | SAR | Interp pts | SAR | |
| 0 | -- | 1812.3 | -- | 1812.3 | |
| 1 | 14 | 483.15 | 14 | 483.15 | |
| 2 | 4,14 | 14.08 | 4,14 | 14.08 | |
| 3 | 4,18 | 8.886 | 4,18 | 8.886 | |
| 4 | 8,18 | 8.870 | 10,18 | 9.275 | **INCREASE!** |
| 5 | | | 8,18 | 8.870 | |

---

[12] This rule was advocated by others seeking to improve the performance of the BR algorithm by reducing the computational load when determining pivots. It requires that each pivot be computed as a weighted median of all positive and negative finite ratios of elements in **R** (except those two elements pertaining to intercept and slope), divided by the corresponding elements in the vector chosen to enter the basis (i.e., the non-zero pivotal column elements), weighted by the absolute value of those same elements in the vector chosen to enter the basis.

More troubling than the previous example, the results below for data set L20S6 show that using pivots chosen *exclusively* by the weighted median rule can lead to a sequence of iterations that cycles endlessly, so that the optimal value SAR* is never reached.

| Data set L20S2 | BR pivots | | WM pivots | |
|---|---|---|---|---|
| Iter no. | Interp pts | SAR | Interp pts | SAR |
| 0 | -- | 2153 | -- | 2153 |
| 1 | 15 | 626.59 | 15 | 626.59 |
| 2 | 8,15 | 13.34 | 8,15 | 13.34 |
| 3 | 8,20 | 12.40 | 8,20 | 12.40 |
| 4 | 10,20 | 11.80 | 3,20 | 11.82 |
| 5 | | | 8,20 | 12.40 |
| 6 | SAR* is 11.80 | | 3,20 | 11.82 |
| ⋮ | | | 12.40↔11.82 Cycles! | |

So, why use this weighted median rule for choosing pivots? The answer is that efficient linear-time algorithms are available for computing weighted medians, and for larger data sets the potential savings in execution times when determining pivots by a weighted median rule can be considerable. However, choosing pivots this way requires careful implementation because (a) it can sometimes require more iterations than BR to reach a solution, (b) it can lead to an increase in SAR values from one iteration to the next, and (c) as was shown above, it can cause cycling to occur. In contrast, the original BR method converges to SAR* in all but extremely rare instances (and ultimately due to limitations in floating-point representations and arithmetic operations).

2.8 Computing weighted L1 fits  When not all the weights are set to 1, just a modest algorithmic change is required to minimizing the SAR defined in expression (1). Illustrating this change by referring to the full initial Simplex tableau in Section 2.1, the costs of 1 appearing in both the left-most column and the upper row would all need to be replaced by $w_1$, $w_2$,..., $w_5$ (or $w_m$ in general). The marginal costs of each pair $u_i$ and $v_i$ would then sum to $-2w_i$, rather than to $-2$ (as in the unweighted[13] case), and the SAR initial value of (1+1+2+3+2) would change to ($w_1+w_2+2w_3+3w_4+2w_5$). Weights are always chosen to be positive real numbers, and they are sometimes normalized to sum to 1. When they are all integers, a weighted fit can be obtained from the original BR code by repeating the $i^{th}$ row $w_i$ times at setup, for all i = 1,2,...,m. This is obviously inefficient computationally, but it can serve to verify through test examples that correct coding changes are made to accommodate all values of $w_i$ that differ from 1.

Data set L5S3 provides an example of a weighted L1 straight-line approximation problem. The left half of the results shown in the figure on the next page contains the initial and final tableaux for the unweighted fit to L5S3, and it is clear from the final marginal costs that the solution d = 83.1+2.05t is unique and that it does not interpolate the last data point (7, 97.8).

---

[13] When all the $w_i$ = 1 we refer to this as the unweighted case.

| Data set L5S3 | Unweighted L1 fit | | | | | Weighted L1 fit | | | | |
|---|---|---|---|---|---|---|---|---|---|---|
| $w_i$ | At beginning of iteration | | | | | $w_i$ | At beginning of iteration | | | |
| 1 | 1 | 3 | 90.4 | 3 | | 1 | 1 | 3 | 90.4 | 3 |
| 1 | 1 | 4 | 91.3 | 4 | | 1 | 1 | 4 | 91.3 | 4 |
| 1 | 1 | 5 | 92.9 | 5 | | 1 | 1 | 5 | 92.9 | 5 |
| 1 | 1 | 6 | 95.4 | 6 | | 1 | 1 | 6 | 95.4 | 6 |
| 1 | 1 | 7 | 97.8 | 7 | | 2 | 1 | 7 | 97.8 | 7 |
| | 5 | 25 | 467.8 | | | | 6 | 32 | 565.6 | |
| | 1 | 2 | | | | | 1 | 2 | | |
| $w_i$ | Final Results | | | | | $w_i$ | Final Results | | | |
| 1 | -0.5 | 0.5 | 2.05 | 2 | | 1 | -0.33333 | 0.33333 | 2.1667 | 2 |
| 1 | 3 | -2 | 83.1 | 1 | | 1 | 2.33333 | -1.33333 | 82.633 | 1 |
| 1 | 0.5 | 0.5 | 0.45 | -5 | | 1 | 0.66667 | 0.33333 | 0.5667 | -5 |
| 1 | -1.5 | 0.5 | 1.15 | 3 | | 1 | -1.33333 | 0.33333 | 1.2667 | 3 |
| 1 | 0.5 | -1.5 | 0.35 | 7 | | 2 | 0.33333 | 0.66667 | 0.2333 | -6 |
| | -1.5 | -1.5 | 1.95 | | | | -1.33333 | -0.66667 | 2.0667 | |
| | 4 | 6 | | | | | 4 | 7 | | |

**Actual CPI value at t = 8 is 100**

L1 prediction at t = 8:   83.1+2.05*8 = 99.50 (unweighted)  &   82.63333+2.166667*8 = 99.97 (weighted)
L2 prediction at t = 8:   84.11+1.89*8 = 99.23 (unweighted)  &   83.88+1.9475*8 = 99.46 (weighted)

Viewing this solution as a model to predict the next CPI value, based on the previous five annual values in L5S3, it seems reasonable to put more emphasis on the last data point by increasing its weight from 1 to 2, say. The right half of the figure shows the resulting initial and final tableaux; the marginal costs and SAR values now must incorporate the weight of 2, and this unique solution d = 82.63333+2.166667t does indeed interpolate the last data point. These L1 approximations predict that at t = 8 the CPI will be 99.50 when the fit is unweighted, and 99.97 when it is weighted, whereas its correct value at t = 8 is 100. Clearly, emphasising the contribution of the most recent point in L5S3 through a weighted L1 fit has improved the prediction for the next CPI value beyond the current data set[14]. Interestingly, the least-squares predictions for this example are not as accurate as those from L1. Weighting does not force L2 fits to interpolate given data points exactly, although a weight of 100 on the data point (7, 97.8) would improve the L2 predicted value at t = 8 to 99.83. Despite intuitive beliefs to the contrary, methods that best fit available data do not *always* provide more accurate predictions!

---

[14] Using a model to predict future values based on previous known values in a time series is the primary role of *time series forecasting*. The above example involves single-step forecasts using just simple *extrapolation* to t = 8 of the four fitted lines. There are many different forecasting algorithms available both for single-step and multi-step forecasting (where, directly and/or recursively, predictions are made for values at two or more consecutive times immediately beyond the known sample). Perhaps there is still room for a univariate L1 forecasting algorithm!

## 2.9 Adding a data point to a set without changing $a_1^*$ and $a_2^*$

Since a best L1 straight-line fit balances the numbers of positive P and negative N residuals so that $|P - N| \leq 2$ (and when m is odd $|P - N| \leq 1$), *without regard to their magnitudes*, this suggests that sometimes a new point could be appended to a given data set without causing the current parameter values $a_1^*$ and $a_2^*$ to change. As can be verified from the second triangular figure in Section 2.4, there are 12 pairs of data sets among the 171 CPI test cases for which the value of SAR* is unchanged from one data set to the one immediately below it. Does this mean that the parameter values for those pairs are unchanged? The answer is sometimes Yes and sometimes No, as is shown in the next two figures using three vertical pairs of data sets from that same triangular summary.

| Data set L4S7 | | ① | | | | Data set L4S4 | | ③ | | |
|---|---|---|---|---|---|---|---|---|---|---|
| t | d | Final Results | | | | t | d | Final Results | | |
| 7 | 97.8 | -0.33333 | -0.33333 | 2.3 | 2 | 4 | 91.3 | -0.5 | 0.5 | 2.05 | 2 |
| 8 | 100 | 3.33333 | 2.33333 | 81.7 | 1 | 5 | 92.9 | 3 | -2 | 83.1 | 1 |
| 9 | 102.8 | 0.66667 | -0.33333 | 0.1 | -4 | 6 | 95.4 | 0.5 | 0.5 | 0.45 | -4 |
| 10 | 104.7 | -0.33333 | 0.66667 | 0.4 | 5 | 7 | 97.8 | 0.5 | -1.5 | 0.35 | 6 |
|  |  | -0.66667 | -0.66667 | 0.5 |  |  |  | 0 | -2 | 0.8 |  |
|  |  | 3 | -6 |  |  |  |  | 3 | 5 |  |  |
| Data set L5S7 | | ② | | | | Data set L5S4 | | ④ | | |
| t | d | Final Results | | | | t | d | Final Results | | |
| 7 | 97.8 | -0.25 | 0.25 | 2.3 | 2 | 4 | 91.3 | -0.5 | -0.5 | 2.3 | 2 |
| 8 | 100 | 1.75 | -2.75 | 81.7 | 1 | 5 | 92.9 | 3 | 4 | 81.6 | 1 |
| 9 | 102.8 | 0.75 | 0.25 | 0 | 6 | 6 | 95.4 | -1 | -2 | 0.5 | 3 |
| 10 | 104.7 | 0.5 | 0.5 | 0.4 | 5 | 7 | 97.8 | 0.5 | -0.5 | 0.1 | 6 |
| 11 | 107 | -0.25 | -0.75 | 0.1 | -4 | 8 | 100 | 0.5 | 1.5 | 0.2 | -4 |
|  |  | 0 | -1 | 0.5 |  |  |  | -1 | -2 | 0.8 |  |
|  |  | -7 | -3 |  |  |  |  | -7 | 5 |  |  |

Tables 1 & 2 show that the unique solution for data set L4S7 is also the unique solution to data set L5S7 (which has the next data point in time appended to L4S7); SAR* = 0.5 for both data sets. The solution to data set L4S4 given in Table 3 has a SAR value of 1.3 when applied to L5S4, so it is not a solution for L5S4 because its Table 4 shows that SAR* = 0.8 when d = 81.6 + 2.3t. A second basic L1 solution for L5S4 is d = 82.4 + 2.2t, and because both basic L1 solutions (since m is odd there are no other basic L1 solutions) fit the last point at t = 8 exactly, they are also both solutions to L4S4 with SAR* = 0.8. Since they interpolate only one point in L4S4 they are not basic L1 solutions for L4S4, so there is no chance that an LP Simplex calculation would uncover them when processing data set L4S4 directly. Consequently, without advance knowledge of the two basic L1 solutions to L5S4, there is no practical computational path from Table 3 to Table 4.

| Data set L5S8 | | | ⑤ | | |
|---|---|---|---|---|---|
| t | d | | Final Results | | |
| 8 | 100 | -0.5 | -0.5 | 2.2 | 2 |
| 9 | 102.8 | 5 | 6 | 82.7 | 1 |
| 10 | 104.7 | 1 | 2 | 0.3 | -3 |
| 11 | 107 | 0.5 | -0.5 | 0.1 | 6 |
| 12 | 109.1 | -0.5 | -1.5 | 0.3 | 4 |
| | | 0 | -1 | 0.7 | |
| | | -7 | 5 | | |
| Data set L6S8 | | | ⑥ | | |
| t | d | | Final Results | | |
| 8 | 100 | -0.33333 | 0.33333 | 2.26667 | 2 |
| 9 | 102.8 | 3.33333 | -4.33333 | 82.0333 | 1 |
| 10 | 104.7 | 0.33333 | 0.66667 | 0.03333 | 6 |
| 11 | 107 | 0.66667 | -1.66667 | 0.16667 | -3 |
| 12 | 109.1 | -0.66667 | -0.33333 | 0.13333 | -7 |
| 13 | 111.5 | -0.33333 | 1.33333 | 0.36667 | 4 |
| | | -1 | -1 | 0.7 | |
| | | -8 | -5 | | |

For Table 5, neither the basic L1 solution shown, nor the second one of d = 81.7 + 2.3t, provides a solution to the expanded data set L6S8, and although its unique solution in Table 6 has SAR* = 0.7, it would be hidden from any practical standpoint when processing data set L5S8.

The largest vertical pair in the triangular summary with identical SAR* values, L17S1 and L18S1, has a unique solution d = 83.9 + 2.1. As also occurs with the three pairs on this and the previous page, that solution interpolates the last point of the larger data set, so SAR* is unchanged for the smaller set (it does not contain that last point).

So, the question remains: is there a shortcut available for obtaining an L1 solution from a previous fit to the *same* data set - apart from one point being added or perhaps deleted?

## 2.10 Updating and downdating a solution

Techniques for updating a general linear programming solution are described in many LP text books, although downdating (i.e., deleting a row or column of a Simplex tableau) receives less attention. In [6], algorithms are given for revising an L1 solution to a general overdetermined system of m linear equations in n unknown parameters after a column or a row of its (m x n) BR condensed tableau is inserted or deleted, or its right-hand side vector **R** is changed. Fortran code for these algorithms is provided in [6], together with appropriate coding modifications to Algorithm 478 [3]. The reasons stated by the authors for utilizing the BR code while developing their algorithms are that it was faster and/or more accurate than other available L1 algorithms, and because BR *does not require the Simplex tableaux to be of full rank* (an important attribute, since modifications made by their algorithms may result legitimately in rank deficiency). Based on their computational experience with test data sets for which m ≤ 200 and 4 ≤ n ≤ 20, the authors note in [6] that impressive gains in speed were attained when compared to solving each new data set from scratch. However, all their reported computational tests involved initial tableaux with the columns and **R** filled with randomly generated data, apart from each first column consisting of all 1's. (My experience is that algorithm evaluations made from tests conducted primarily with random data can differ from conclusions gained using real data drawn from typical applications of those algorithms.)

While adding or deleting columns is of no relevance for straight line fitting, the ability to add or delete rows allows revisions to be made in an L1 solution to accommodate changes to a current data set's number m of data points, and their values. There is no discussion or computational evidence in [6] concerning L1 straight line fitting, although there is no reason why their algorithms could not be effective when incorporated in a BRSL algorithm implementation.

### 2.11 Changes in d values
Another intriguing characteristic of L1 fitting (applicable in general, not just for straight-line fits) can be seen from the next figure.

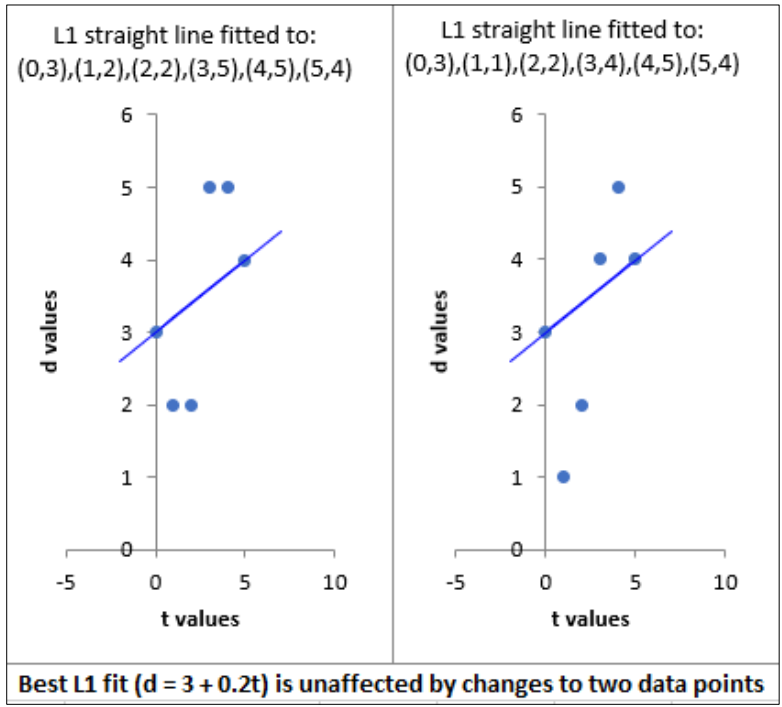

Here, two of the d values from the data set shown in the left panel have been changed, as shown in the right panel, but the unique best L1 solution remains unchanged, as d = 3 + 0.2t. This would have already been known without calculating the fit to the perturbed data set, because these two changes to the data shown in the left panel do not cause positive residuals to become negative, or vice versa. Indeed, if we altered the left data set by changing the two values of 5 to 100, and the two values of 2 to 0, or if we altered the right data set by adding 100 to each of the d values for t = 3 and t = 4 and replaced (1,1) and (2,2) with (1,-1) and (2, -15), in either case the best L1 fit would still be d = 3 + 0.2t. Unlike L2 (least-squares) fits, which are sensitive to the *size* of its residuals, L1 fits are sensitive to the *signs* of their residuals. Consequently, when processing real data sets that might contain outliers, L1 fits can provide a different lens through which to view and assess data. Most LP textbooks contain discussions on methods for conducting *sensitivity analyses*, whereby changes to an optimal LP solution due to changes in the input data can be determined. In the case of L1 fitting, it is obvious in the figure above that non-interpolated d values above the fitted line could be increased indefinitely, and those below decreased indefinitely, without any change to the optimal values $a_1^*$ or $a_2^*$.

### 2.11 Further revelations from a real time series
There are sources for downloading free real data sets that can be used for algorithm testing. One reliable source is the U.S. Geological Survey's National Water Information System; the following remarks and results concern a USGS time series consisting of mean daily streamflow (i.e., rate of water discharge occurring in a natural channel), measured in cubic feet per second, and downloaded from the link Streamflow Time Series. This 3652-point ten-year data set for the period 2001 - 2010 has minimum and maximum values of 20 and 5380, but half its values are less than or equal to 132. Indeed, from the first three bars of its histogram shown next it can be confirmed that 72% of these streamflow values do not exceed 288, although there are more than a few days in the remaining 28% where the streamflow values differ substantially from most other values (e.g., there are more than 200 values greater than 1000).

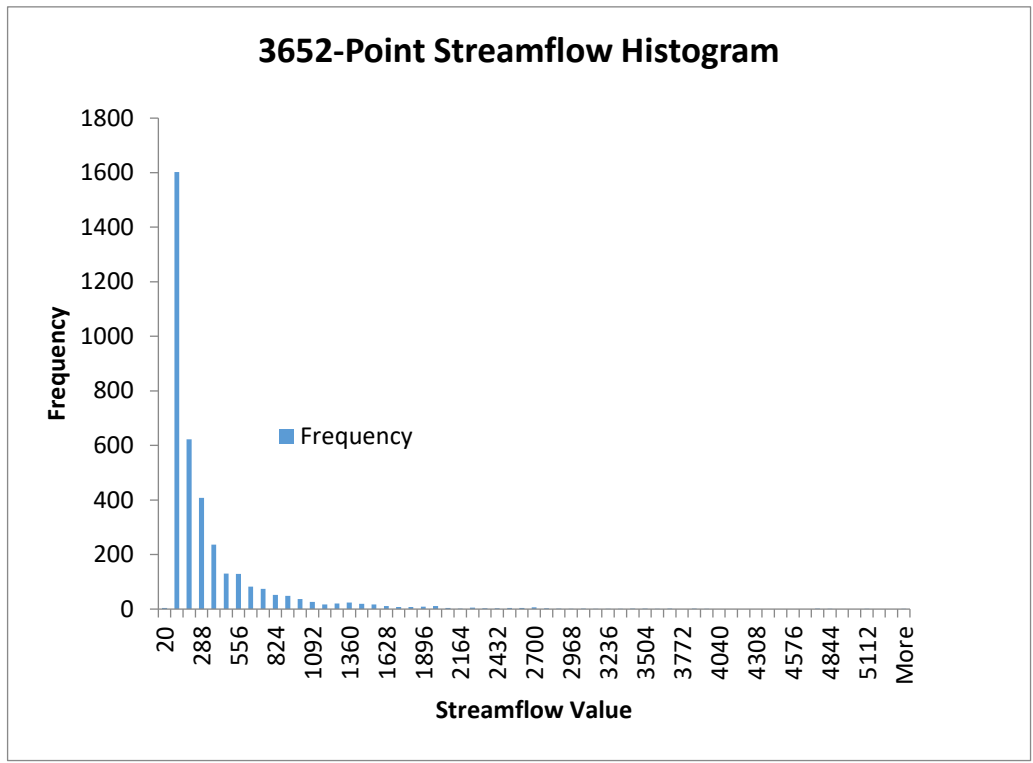

Values that lie an abnormal distance from the mass of a data set are commonly referred to as outliers; these may sometimes be due to experimental error, recording mistakes, or outright blunders. But all these USGS data values are quality-assured, being averages of many streamflow measurements recorded each day; their extreme variability and observed periodicity are undoubtedly due to seasonal changes in weather. Whatever their provenance, outliers can cause serious problems in statistical analyses, and analysts must decide how to handle such points. Just rejecting them may not be an option (particularly for a time series whose successive values all need to be recorded at equal time intervals), so algorithms that are not unduly influenced by outliers (so-called *robust* methods) can be of real value in practice.

It is well known that *least-squares (L2) solutions are more influenced by outliers than least-absolute-residuals (L1) solutions*, and this can be seen in the next diagram showing two hydrographs, one for all ten years of streamflow data and the other for just the first year, both overlaid with their L1 (blue line) solution and L2 (red line) solution. The positioning of the L2 lines has been influenced by the presence of outliers in both the 2001-2010 data set and in its 2001 portion.

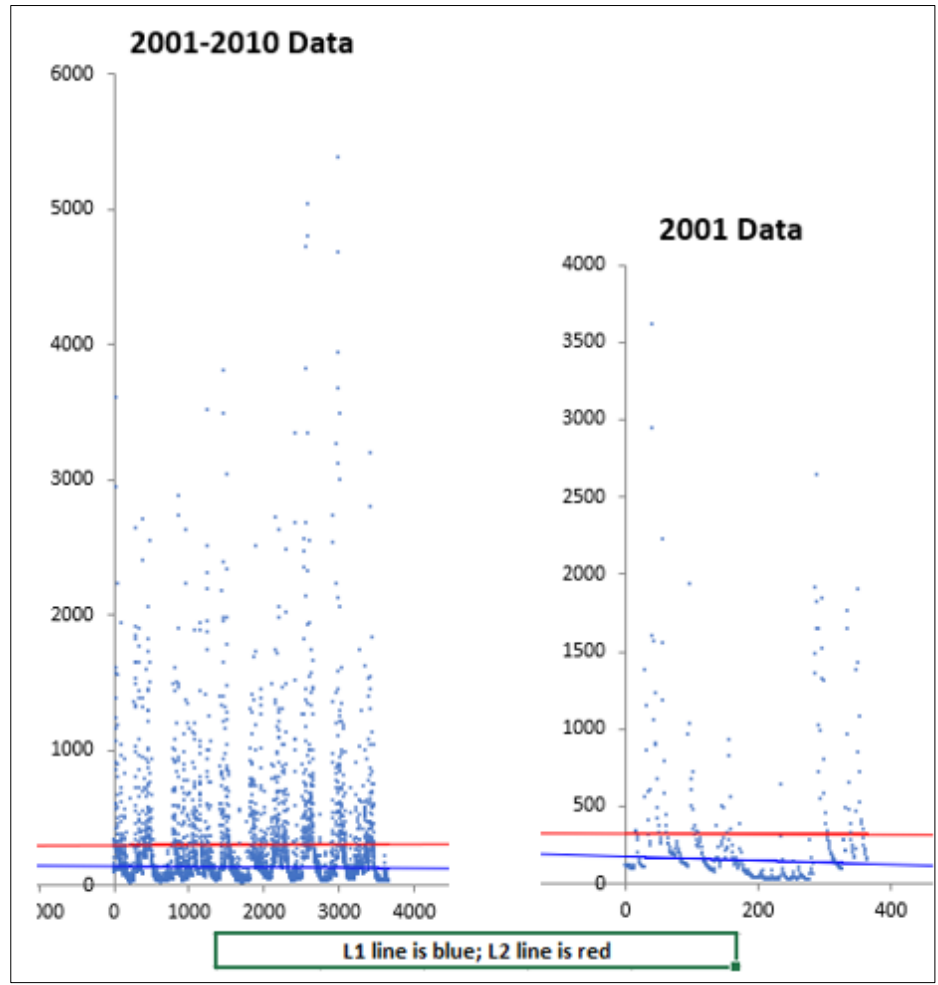

Computations with this streamflow data containing outliers proved to be of more general interest relative to the central theme of this report. It transpired that for either data set (a) starting from the L2 solution did *not* lead to fewer iterations than when the original BR algorithm was employed from a cold start, and (b) the original BR algorithm converged in *fewer iterations* than it did when pivots were chosen using weighted medians. For example, with the 365-point 2001 data set, 6 iterations were required when using weighted median pivots versus just 4 iterations with BR pivots; the unique L1 solution here is d = 173.7070707 - 0.116161616t.

A further observation from the hydrographs is that the L2 solutions for both the one-year and ten-year time series have noticeably more negative residuals than positive ones (75.6% are negative for 2001 data and 73.45% are negative for the 2001-2010 data). It is not surprising then that these two L2 solutions are unsuitable as L1 initial estimates, since it would be known in advance of any computations that the difference in number of positive and negative L1 residuals could not exceed 1; it is 1 for the 2001 data and 0 for the 2001-2010 data.

For the record, the mean streamflow value for the 3652-point data set is 301.26 and the L2 solution is d = 299.7970941 + 0.000800556t. Its unique L1 solution, obtained using the original

BR method in just 5 iterations, from a cold start with 3652 rows of data in the initial tableau, is d = 137.3693143 + 0.002715547t; this data set has a median value of 132, as noted previously.

## 2.12 Five methods to explore

Even for the two-parameter case of fitting straight lines to data, there are several methods for choosing the pivots in each iteration. Five such methods are given below and illustrated by computations with the two CPI data sets L11S3 and L9S4. The five methods are as follows:

(i) The pivot column is the one with largest positive marginal cost (in the event of a tie, it's the first one encountered), and the *BR method* is used to bypass successive potential pivots (in non-decreasing order of ratios of elements of **R** divided by the corresponding *positive* elements in the pivot column) until the first such element is encountered that causes the pivot column's successively smaller marginal cost to become negative, each reduction being by twice the size of the potential pivot (the bypassed element in the pivot column).

(ii) The pivot column is chosen as in (i) above, and the weighted median *WM method* is used to select the pivot row by determining the weighted median of all positive and negative finite ratios of elements in **R** (except those in the two rows pertaining to intercept and slope) each divided by its corresponding element in the pivot column, weighted by the absolute value of that pivot column element.

(iii) The pivotal column is chosen as in (i) above, and for the first two iterations the pivot is determined via the WM method described in (ii) above; these first two pivots are also BR pivots, but calculated more efficiently than in (i). *After completing two iterations, the current solution and residuals are immediately available and are then used as starting values in a second application of this two-iteration scheme, and so.* This scheme avoids occasional increases in SAR that can occur with (ii) above, while benefitting from faster (weighted median) pivot determinations than in (i).

(iv) The least-squares or L2 solution is first determined (the closed form formulas given as footnote 6 in Section 1.6 can be used), the *L2 residuals are next computed, and then the BR method (i) above is used to calculate an L1 fit to the L2 residuals*. Upon convergence these L1 intercept and slope values are added to the L2 intercept and slope values to obtain an L1 solution to the original data. Whenever the L2 solution and residuals are required anyway, they provide starting values at no additional cost for the L1 computations.

(v) Departing from the previous approaches, *the first pivot column would be the intercept column (Col 1)*, since the pivot then corresponds to just a median of the elements in **R**. The second pivot column would be the slope column (Col 2), and its pivot would correspond to a weighted median (as described in (ii) above). By applying *this strategy for choosing pivotal columns, with pivots chosen using medians alternating with weighted medians, to method (iii)* (with or without an L2 start), there could be scope for further computational savings.

For data set L11S3 methods (i) and (ii) converge in just two iterations, and so the BR and WM pivots are identical here. Method (iii) is not required, since for its first two iterations it is just method (ii). The fast start provided by method (iv) is not necessary, since the L1 fit has been produced from a cold start in two iterations. (For the record, method (iv) also takes two iterations to produce the L1 fit to the L2 residuals, and adding this solution to the intercept and slope of the L2 fit gives a best L1 straight line fit of d = 82.0333 + 2.26667t to data set L11S3.)

The first figure below shows the (condensed) Simplex tableaux for the two-iteration solution via the BR (and WM) method. The order in which the intercept or slope vector enters the basis first is determined by the larger of their initial marginal costs (88 vs 11), and the remaining vector enters in the second iteration; the SAR value is reduced from 1102.9 to 280.18 to 3.26667.

Data set L11S3   Methods (i) & (ii): BR (and WM) converges in 2 iterations

| At beginning of iteration | | | | After first iteration | | | | Final Results | | | |
|---|---|---|---|---|---|---|---|---|---|---|---|
| 1 | 3 | 90.4 | 3 | 0.1 | 0.1 | 10.47 | 2 | -0.33333 | -0.33333 | 2.26667 | 2 |
| 1 | 4 | 91.3 | 4 | 0.6 | -0.4 | 49.42 | 4 | 3.33333 | 4.33333 | 82.0333 | 1 |
| 1 | 5 | 92.9 | 5 | 0.5 | -0.5 | 40.55 | 5 | 1.66667 | 2.66667 | 0.46667 | -5 |
| 1 | 6 | 95.4 | 6 | 0.4 | -0.6 | 32.58 | 6 | 1.33333 | 2.33333 | 0.23333 | -6 |
| 1 | 7 | 97.8 | 7 | 0.3 | -0.7 | 24.51 | 7 | 1 | 2 | 0.1 | -7 |
| 1 | 8 | 100 | 8 | 0.2 | -0.8 | 16.24 | 8 | 0.66667 | 1.66667 | 0.16667 | -8 |
| 1 | 9 | 102.8 | 9 | 0.1 | -0.9 | 8.57 | 9 | -0.33333 | -1.33333 | 0.36667 | 9 |
| 1 | 10 | 104.7 | 10 | 0.7 | -0.3 | 58.99 | 3 | -2.33333 | -3.33333 | 1.56667 | 3 |
| 1 | 11 | 107 | 11 | 0.1 | 1.1 | 8.17 | -11 | 0.33333 | -0.66667 | 0.03333 | 11 |
| 1 | 12 | 109.1 | 12 | 0.2 | 1.2 | 16.54 | -12 | -0.66667 | 0.33333 | 0.13333 | -12 |
| 1 | 13 | 111.5 | 13 | 0.3 | 1.3 | 24.61 | -13 | -2 | -3 | 0.2 | 4 |
| 11 | 88 | 1102.9 | | 3.4 | -1.6 | 280.18 | | -1.33333 | -0.33333 | 3.26667 | |
| 1 | 2 | | | 1 | 10 | | | -13 | 10 | | |

Data set L11S3   First iteration of method (v) reduces SAR more effectively

| At beginning of iteration | | | | After first iteration | | | |
|---|---|---|---|---|---|---|---|
| 1 | 3 | 90.4 | 3 | 1 | 8 | 100 | 1 |
| 1 | 4 | 91.3 | 4 | 1 | 4 | 8.7 | -4 |
| 1 | 5 | 92.9 | 5 | 1 | 3 | 7.1 | -5 |
| 1 | 6 | 95.4 | 6 | 1 | 2 | 4.6 | -6 |
| 1 | 7 | 97.8 | 7 | 1 | 1 | 2.2 | -7 |
| 1 | 8 | 100 | 8 | 1 | 5 | 9.6 | -3 |
| 1 | 9 | 102.8 | 9 | -1 | 1 | 2.8 | 9 |
| 1 | 10 | 104.7 | 10 | -1 | 2 | 4.7 | 10 |
| 1 | 11 | 107 | 11 | -1 | 3 | 7 | 11 |
| 1 | 12 | 109.1 | 12 | -1 | 4 | 9.1 | 12 |
| 1 | 13 | 111.5 | 13 | -1 | 5 | 11.5 | 13 |
| 11 | 88 | 1102.9 | | -1 | 30 | 67.3 | |
| 1 | 2 | | | 8 | 2 | | |

Turning now to the strategy embodied in method (v), observe how the BR method (i) and the WM method (ii) process data set L11S3 when the intercept vector enters the basis in the first iteration. The figure to the left shows that the SAR value is reduced from 1102.9 to 67.3 in the first iteration - a more impressive reduction than is made by the other methods (just to 280.18); also, the pivot is located opposite the *median* of the RHS d-values (no weighted median computation required). This gain is short-lived because BR now takes four iterations to converge. The next figure displays successes (methods (i) and (ii) both converge in two iterations, being the minimum number of iterations possible), disappointment (method (v) using BR pivots takes twice as many iterations), and outright failure when method (v) uses WM pivots. This failure in method (v) occurs in its fourth iteration when WM pivoting replaces the interpolation point of t = 6 with t = 11 (rather than t = 10, as with BR), and then cycling occurs!

| t | d | Data set L11S3 | | Methods (i) & (ii) | | Col 1 first | | Col 1 first | |
|---|---|---|---|---|---|---|---|---|---|
| | | | | BR & WM pivots | | Method (i) | | Method (ii) | |
| 3 | 90.4 | | Iter no. | Interp pts | SAR | Interp pts | SAR | Interp pts | SAR |
| 4 | 91.3 | | 0 | -- | 1102.9 | -- | 1102.9 | -- | 1102.9 |
| 5 | 92.9 | | 1 | 10 | 280.18 | 8 | 67.3 | 8 | 67.3 |
| 6 | 95.4 | | 2 | 10,13 | 3.26667 | 6,8 | 3.5 | 6,8 | 3.5 |
| 7 | 97.8 | | 3 | | | 6,13 | 3.5 | 6,13 | 3.5 |
| 8 | 100 | | 4 | SAR* is 3.266667 | | 10,13 | 3.26667 | 11,13 | 3.35 |
| 9 | 102.8 | | 5 | | | | | 7,13 | 3.28333 |
| 10 | 104.7 | | 6 | | | SAR* is 3.266667 | | 11,13 | 3.35 |
| 11 | 107 | | 7 | | | | | 7,13 | 3.28333 |
| 12 | 109.1 | | ⋮ | | | | | 3.35↔3.283333 Cycles! | |
| 13 | 111.5 | | | | | | | | |

Fortunately, this type of disastrous behavior can be avoided by *not* using the WM method (ii) *blindly*. As shown below, the two-iteration repetitive method (iii) with the first pivotal column always being Col 1 avoided cycling. However, convergence did require a total of six Simplex iterations, whether maximum marginal costs determine pivotal columns or Col 1 is used first.

| Initially (pivots in boxes) | | | | After one iteration | | | | After two iterations | | | | Residuals (input for iteration 3) | | |
|---|---|---|---|---|---|---|---|---|---|---|---|---|---|---|
| 1 | 3 | 90.4 | 3 | 1 | 8 | 100 | 1 | 3 | -4 | 81.6 | 1 | 1.9 | 3 | |
| 1 | 4 | 91.3 | 4 | 1 | 4 | 8.7 | -4 | -0.5 | 0.5 | 2.3 | 2 | 0.5 | 4 | |
| 1 | 5 | 92.9 | 5 | 1 | 3 | 7.1 | -5 | 0.5 | -1.5 | 0.2 | -5 | -0.2 | 5 | |
| 1 | 6 | 95.4 | 6 | 1 | [2] | 4.6 | -6 | -1 | 2 | 0.5 | 4 | 0 | 6 | |
| 1 | 7 | 97.8 | 7 | 1 | 1 | 2.2 | -7 | 0.5 | 0.5 | 0.1 | 7 | 0.1 | 7 | |
| 1 | [8] | 100 | 8 | 1 | 5 | 9.6 | -3 | -1.5 | 2.5 | 1.9 | 3 | 0 | 8 | |
| 1 | 9 | 102.8 | 9 | -1 | 1 | 2.8 | 9 | 1.5 | -0.5 | 0.5 | 9 | 0.5 | 9 | |
| 1 | 10 | 104.7 | 10 | -1 | 2 | 4.7 | 10 | 2 | -1 | 0.1 | 10 | 0.1 | 10 | |
| 1 | 11 | 107 | 11 | -1 | 3 | 7 | 11 | 2.5 | -1.5 | 0.1 | 11 | 0.1 | 11 | |
| 1 | 12 | 109.1 | 12 | -1 | 4 | 9.1 | 12 | -3 | 2 | 0.1 | -12 | -0.1 | 12 | |
| 1 | 13 | 111.5 | 13 | -1 | 5 | 11.5 | 13 | 3.5 | -2.5 | 0 | 13 | 0 | 13 | |
| 11 | 88 | 1102.9 | | -1 | 30 | 67.3 | | 4 | -1 | 3.5 | | | | |
| 1 | 2 | | | 8 | 2 | | | 8 | -6 | | | | | |

| Initially (for iterations 3 & 4) | | | | After three iterations | | | | After four iterations | | | | Residuals (input for iteration 5) | | |
|---|---|---|---|---|---|---|---|---|---|---|---|---|---|---|
| 1 | 3 | 1.9 | 3 | 1 | -7 | 0.1 | 1 | -2.1667 | 1.16667 | 0.21667 | 1 | 1.73333 | 3 | |
| 1 | 4 | 0.5 | 4 | -1 | 3 | 0.4 | 4 | -0.1667 | 0.16667 | 0.01667 | -2 | 0.35 | 4 | |
| -1 | -5 | 0.2 | -5 | 1 | -2 | 0.3 | -5 | -1.3333 | 0.33333 | 0.33333 | -5 | -0.3333 | 5 | |
| 1 | 6 | 0 | 6 | 1 | -1 | 0.1 | -6 | -1.1667 | 0.16667 | 0.11667 | -6 | -0.1167 | 6 | |
| 1 | [7] | 0.1 | 7 | -1 | 4 | 1.8 | 3 | 1.66667 | -0.6667 | 1.73333 | 3 | 0 | 7 | |
| 1 | 8 | 0 | 8 | 1 | 1 | 0.1 | -8 | -0.8333 | -0.1667 | 0.08333 | -8 | -0.0833 | 8 | |
| 1 | 9 | 0.5 | 9 | -1 | -2 | 0.4 | 9 | 0.66667 | 0.33333 | 0.43333 | 9 | 0.43333 | 9 | |
| 1 | 10 | 0.1 | 10 | -1 | -3 | 0 | 10 | 0.5 | 0.5 | 0.05 | 10 | 0.05 | 10 | |
| 1 | 11 | 0.1 | 11 | -1 | -4 | 0 | 11 | 0.33333 | 0.66667 | 0.06667 | 11 | 0.06667 | 11 | |
| -1 | -12 | 0.1 | -12 | 1 | 5 | 0.2 | -12 | -0.1667 | -0.8333 | 0.11667 | -12 | -0.1167 | 12 | |
| 1 | 13 | 0 | 13 | 1 | [6] | 0.1 | -13 | 1.5 | -0.5 | 0.35 | 4 | 0 | 13 | |
| 11 | 88 | 1102.9 | | -1 | 7 | 3.4 | | 0.16667 | -1.1667 | 3.28333 | | | | |
| 1 | 2 | | | 7 | -2 | | | -7 | -13 | | | | | |

| Initially (for iterations 5 & 6) | | | | After five iterations | | | | Final Results (after six iterations) | | | | | | |
|---|---|---|---|---|---|---|---|---|---|---|---|---|---|---|
| 1 | 3 | 1.73333 | 3 | 1 | -13 | 0 | 1 | -3.3333 | 4.33333 | 0.21667 | 1 | Intercept is | 81.6 | |
| 1 | 4 | 0.35 | 4 | -1 | 9 | 0.35 | 4 | -0.3333 | 0.33333 | 0.01667 | -2 | | + 0.21667 | |
| -1 | -5 | 0.33333 | -5 | 1 | -8 | 0.33333 | -5 | -1.6667 | 2.66667 | 0.46667 | -5 | | + 0.21667 | |
| -1 | -6 | 0.11667 | -6 | 1 | -7 | 0.11667 | -6 | -1.3333 | 2.33333 | 0.23333 | -6 | | = 82.0333 | |
| 1 | 7 | 0 | 7 | -1 | 6 | 0 | 7 | -1 | 2 | 0.1 | -7 | | | |
| -1 | -8 | 0.08333 | -8 | 1 | -5 | 0.08333 | -8 | -0.6667 | 1.66667 | 0.16667 | -8 | Slope is | 2.3 | |
| 1 | 9 | 0.43333 | 9 | -1 | 4 | 0.43333 | 9 | 0.33333 | -1.3333 | 0.36667 | 9 | | - 0.01667 | |
| 1 | 10 | 0.05 | 10 | -1 | [3] | 0.05 | 10 | 2 | -3 | 0.2 | 4 | | - 0.01667 | |
| 1 | 11 | 0.06667 | 11 | -1 | 2 | 0.06667 | 11 | -0.3333 | -0.6667 | 0.03333 | 11 | | = 2.26666 | |
| -1 | -12 | 0.11667 | -12 | 1 | -1 | 0.11667 | -12 | 0.66667 | 0.33333 | 0.13333 | -12 | | | |
| 1 | 13 | 0 | 13 | -1 | 10 | 1.73333 | 3 | 2.33333 | -3.3333 | 1.56667 | 3 | | | |
| 3 | 26 | 3.28333 | | -3 | 13 | 3.28333 | | -0.6667 | -0.3333 | 3.26667 | | | | |
| 1 | 2 | | | 13 | -2 | | | 13 | 10 | | | | | |

Computing L1 Straight-Line Fits to Data (Part 1)                                     December 2019

Experience applying methods (i)-(v) to the data set L9S4 is summarized in the next two figures. Method (i) required five iterations to converge and method (ii) took six (note the fourth WM iteration). Using the L2 solution for a fast start, method (iv) took just two iterations to converge. By forcing Col 1 to be the first pivotal column and then applying BR pivots or WM pivots, method (v) first reduced SAR to 46.2 (compared to 190.867 via methods (i) or (ii)), and it took one iteration less than method (i) and two less than method (ii). But the lower figure exposes a serious flaw in method (iii); its two-iteration repetitions can cause cycling!

| t | d | Data set L9S4 | Method (i) BR pivots | | Method (ii) WM pivots | | Method (iv) BR or WM pivots | | Col 1 first BR or WM pivots | |
|---|---|---|---|---|---|---|---|---|---|---|
| | | Iter no. | Interp pts | SAR | Interp pts | SAR | Interp pts | SAR | Interp pts | SAR |
| 4 | 91.3 | 0 | -- | 901 | -- | 901 | -- | 1.57111 | -- | 901 |
| 5 | 92.9 | 1 | 9 | 190.867 | 9 | 190.867 | 7 | 1.55619 | 8 | 46.2 |
| 6 | 95.4 | 2 | 4,9 | 3.5 | 4,9 | 3.5 | 7,10 | 1.5 | 6,8 | 1.6 |
| 7 | 97.8 | 3 | 4,12 | 1.9 | 4,12 | 1.9 | | | 6,11 | 1.52 |
| 8 | 100 | 4 | 7,12 | 1.62 | 8,12 | 1.65 | | | 7,11 | 1.5 |
| 9 | 102.8 | 5 | 7,10 | 1.5 | 7,12 | 1.62 | | | | |
| 10 | 104.7 | 6 | | | 7,10 | 1.5 | | | | |
| 11 | 107 | | | | | | | | | |
| 12 | 109.1 | | | | | | | | | |

| | Initially | | | After two iterations | | | Residuals (input for iteration 3) | |
|---|---|---|---|---|---|---|---|---|
| 1 | 4 | 91.3 | 3 | -0.2 | 0.2 | 2.3 | 2 | 0 | 3 |
| 1 | 5 | 92.9 | 4 | 1.8 | -0.8 | 82.1 | 1 | -0.7 | 4 |
| 1 | 6 | 95.4 | 5 | 0.6 | 0.4 | 0.5 | -5 | -0.5 | 5 |
| 1 | 7 | 97.8 | 6 | 0.4 | 0.6 | 0.4 | -6 | -0.4 | 6 |
| 1 | 8 | 100 | 7 | 0.2 | 0.8 | 0.5 | -7 | -0.5 | 7 |
| 1 | 9 | 102.8 | 8 | 0.8 | 0.2 | 0.7 | -4 | 0 | 8 |
| 1 | 10 | 104.7 | 9 | -0.2 | 1.2 | 0.4 | -9 | -0.4 | 9 |
| 1 | 11 | 107 | 10 | -0.4 | 1.4 | 0.4 | -10 | -0.4 | 10 |
| 1 | 12 | 109.1 | 11 | -0.6 | 1.6 | 0.6 | -11 | -0.6 | 11 |
| 9 | 72 | 901 | | -0.2 | 5.2 | 3.5 | | | |
| 1 | 2 | | | 3 | 8 | | | | |

| | Initially (for iterations 3 & 4) | | | After four iterations | | | Residuals (input for iteration 5) | |
|---|---|---|---|---|---|---|---|---|
| 1 | 4 | 0 | 3 | -0.2 | 0.2 | 0.04 | -2 | 0.28 | 3 |
| -1 | -5 | 0.7 | -4 | 2.4 | -1.4 | 0.12 | -1 | -0.38 | 4 |
| -1 | -6 | 0.5 | -5 | -1.2 | 0.2 | 0.14 | -5 | -0.14 | 5 |
| -1 | -7 | 0.4 | -6 | -1.4 | 0.4 | 0.38 | -4 | 0 | 6 |
| -1 | -8 | 0.5 | -7 | -0.8 | -0.2 | 0.06 | -7 | -0.06 | 7 |
| 1 | 9 | 0 | 8 | 0.6 | 0.4 | 0.48 | 8 | 0.48 | 8 |
| -1 | -10 | 0.4 | -9 | 0.4 | 0.6 | 0.12 | 9 | 0.12 | 9 |
| -1 | -11 | 0.4 | -10 | 0.2 | 0.8 | 0.16 | 10 | 0.16 | 10 |
| -1 | -12 | 0.6 | -11 | 1.6 | -0.6 | 0.28 | 3 | 0 | 11 |
| -5 | -46 | 3.5 | | -1.6 | 0.6 | 1.62 | | | |
| 1 | 2 | | | -6 | -11 | | | | |

| | Initially (for iterations 5 & 6) | | | After six iterations | | | | |
|---|---|---|---|---|---|---|---|---|
| 1 | 4 | 0.28 | 3 | -0.2 | 0.2 | 0 | 2 | Intercept is | 82.1 |
| -1 | -5 | 0.38 | -4 | 2.2 | -1.2 | 0 | 1 | + | (-0.12) |
| -1 | -6 | 0.14 | -5 | 1.2 | -0.2 | 0.14 | -5 | + | 0 |
| 1 | 7 | 0 | 6 | 1.4 | -0.4 | 0.38 | -4 | = | 81.98 |
| -1 | -8 | 0.06 | -7 | 0.8 | 0.2 | 0.06 | -7 | | |
| 1 | 9 | 0.48 | 8 | -0.6 | -0.4 | 0.48 | 8 | Slope is | 2.3 |
| 1 | 10 | 0.12 | 9 | -0.4 | -0.6 | 0.12 | 9 | + | (-0.04) |
| 1 | 11 | 0.16 | 10 | -0.2 | -0.8 | 0.16 | 10 | + | 0 |
| 1 | 12 | 0 | 11 | -1.6 | 0.6 | 0.28 | 3 | | 2.26 |
| 3 | 34 | 1.62 | | -0.4 | -2.6 | 1.62 | | | |
| 1 | 2 | | | 6 | 11 | Method (iii) cannot progress further towards SAR* = 1.5 | | | |

All is not lost though, as the compromise proposed next in Recommendations can outperform BR in speed of execution, without sacrificing a guarantee of convergence. As a final observation in this Section 2.12, the L1 solution d = 81.7 + 2.3t for data set L9S4 interpolates three of its data points, at t = 6, 9, and 10. This causes a degeneracy in the final Simplex tableau, but inspection of that tableau below shows that this L1 straight-line solution is unique.

| At beginning of iteration | | | | After first iteration | | | | After fourth (final) iteration | | | | Three basic feasible | |
|---|---|---|---|---|---|---|---|---|---|---|---|---|---|
| 1 | 4 | 91.3 | 3 | 1 | 8 | 100 | 1 | -1.75 | 2.75 | 81.7 | 1 | LP solutions, but the | |
| 1 | 5 | 92.9 | 4 | 1 | 3 | 7.1 | -4 | 0.25 | -0.25 | 2.3 | 2 | L1 solution is unique | |
| 1 | 6 | 95.4 | 5 | 1 | 2 | 4.6 | -5 | -0.5 | 1.5 | 0.3 | -4 | | |
| 1 | 7 | 97.8 | 6 | 1 | 1 | 2.2 | -6 | -0.25 | 1.25 | 0.1 | -5 | t values | labels |
| 1 | 8 | 100 | 7 | 1 | 4 | 8.7 | -3 | 0.75 | -1.75 | 0.4 | 3 | 7,10 | 6,9 |
| 1 | 9 | 102.8 | 8 | -1 | 1 | 2.8 | 8 | -0.5 | -0.5 | 0.4 | 8 | 7,11 | 6,10 |
| 1 | 10 | 104.7 | 9 | -1 | 2 | 4.7 | 9 | -0.75 | -0.25 | 0 | 9 | 10,11 | 9,10 |
| 1 | 11 | 107 | 10 | -1 | 3 | 7 | 10 | 0.25 | 0.75 | 0.1 | -7 | | |
| 1 | 12 | 109.1 | 11 | -1 | 4 | 9.1 | 11 | 1.25 | -0.25 | 0.2 | -11 | | |
| 9 | 72 | 901 | 0 | -1 | 20 | 46.2 | | -0.75 | -0.25 | 1.5 | | | |
| 1 | 2 | 0 | 0 | 7 | 2 | | | 10 | 6 | —degeneracy | | | |

# 3 Recommendations

Emphasis has been placed throughout this report on how well a certain rule or process reduces the SAR value from one iteration to the next iteration. In linear programming terminology SAR is the objective function, and one must always be aware that the Simplex method cannot look ahead past its next iteration, and that a current best step might lead to a poor solution path overall. However, based on its performance (stretching back to the 1950's) with thousands of real-life LP applications involving many millions of different data sets, the Simplex method has proved to be an extremely popular, efficient, and reliable computational algorithm.

When determination of an L1 best fit by a straight line is looked upon as simply a search for a best pair of interpolation points, Simplex-type methods perform this task remarkably well. For example, the seven CPI data groups with the longer data sets (i.e., m = 15, 16, …, 21) provide a total of 28 separate data sets (L15S1, L15S2, …, L21S1), each one with between 105 to 210 possible best interpolation pairs. Nevertheless, the algorithm recommended here takes only 2, 3, 4, or 5 iterations to converge for any of those 28 cases, and in only 8 cases are 5 iterations needed. Also, in Section 2.11 it was noted that the BR method (i) took just 4 iterations to fit an L1 straight line to the 2001 annual segment of 365 Streamflow data points. Quite impressive!

In spite of gains made when the Simplex method was adapted via the BR method (i) to deal efficiently with the LP formulation (2) posed in Section 2.1, the purpose of this report has been to raise awareness of some additional algorithmic improvements that should render an updated BRSL method more suitable than the original BR for L1 straight-line fitting to today's big data sets. Based on computational experiments with different BRSL variations applied to various artificial and real test data sets (a very small sample of which appear in Section 2.12), my recommended approach to developing code embodying a computationally efficient and trustworthy Simplex-based method for L1 line fitting is as follows.

    A. The core engine is method (ii), using the largest marginal cost to determine each pivotal column, and with each pivot determined by the weighted median rule given in Section 2.7 as footnote 12. At the end of any iteration the latest value of SAR must now be retained for comparison with its value at completion of the next iteration.

    B. To restrict occasional increases in SAR as iterations proceed, and to safeguard against the possibility of cycling, a switch to the BR pivot selection rule of method (i) must be enforced for the next iteration immediately following any iteration in which the value of SAR does not decrease. Subsequent pivots continue to be determined as weighted medians, unless another SAR non-decrease occurs, in which case another BR pivot intervention is made, and so on until convergence is achieved.

    C. The code should provide built-in provisions for options to
        (a) begin the iterations from a cold start, or with an internally generated L2 solution and residuals, or with a user-supplied trial solution and residuals,
        (b) calculate weighted L1 fits using user-supplied weights, and
        (c) scale the columns to
- balance the input data (as in footnote 11 of Section 2.5), or
- force either the intercept or the slope column to enter the basis first.

After a pivot switch is made, as called for in step B of this BRSL algorithm, the transformation of the current tableau can proceed seamlessly with the BR pivot. Then the next tableau can be transformed seamlessly again, almost always by using a weighted median pivot; the exception being when no decrease in SAR results after using a BR pivot, and so BR iterations must continue until a decrease in SAR permits a return to using weighted median pivots again.

Two examples of rescues from cycling by switching to BR pivots from WM pivots occurred following the sixth iteration shown in the right most column of the top figure on page 21 (where a SAR value *increased* to 3.35 from 3.28333), and following the third two-iteration pass shown in the lower figure on page 22 (where SAR did *not decrease* after completing the third pass). For both cases solutions were found after just one extra iteration using BR pivoting, so no switching back to WM pivoting was required (the computational details are not shown here).

# 4 Final remarks

It is important to note, as was done previously in Section 2.10, that development and testing of algorithms should always involve the use of real data, rather than just contrived or artificial data (e.g., arrays filled with random numbers). Including real and larger data sets is essential, and it sometimes leads to informative and unexpected outcomes that can expose previously held assertions as being more apparent than real. It may well be in marketing that perception is more important than reality, but not so in computational science.

In point of fact this report may have created the perception that using weighted median pivots does not provide a sufficient advantage over choosing pivots via the simple rule used in BR method (i), particularly in view of occasional WM glitches (even though step B above shows how to resolve such lapses). But tests with large real data sets show that it is not uncommon for method (i) to involve approximately m or so total bypasses in only a handful of iterations. For example, a total of 1,636,584 bypasses were used in the 7 iterations taken to fit a certain logarithmic test data set for which m = 1,638,401. In contrast, method (ii) took 8 iterations using pivots determined as weighted medians, so no bypasses were required, and the saving in computation time was huge. When a comparison was made using just 401 equally spaced data points sampled from the 1,636,584 points over the same time period, method (i) took 5 iterations and method (ii) took 7 iterations, and for this much smaller data set the two computation times were almost equal. Clearly, BR versus WM timing comparisons using just a few hundred points from this large data set could lead to the wrong impression about the value of WM pivoting when large data sets need to be fitted.

# Acknowledgements

My former company colleagues Michael Dunham-Wilkie and Ruining Wu provided valuable assistance with test software development and earlier numerical experiments. The worked examples, conclusions, and any errors occurring in this report are my responsibility; please email any comments, criticisms, or corrections directly to me. I am also grateful to Professor Yvonne Coady for including this report in a directed studies course for some of her students at the University of Victoria. The references below are just to provide sources of further information for students and others assigned the task of writing code for the BRSL algorithm outlined in this report. It is not intended to be an adequate list of references for more in-depth studies of L1 approximation algorithms and their various implementations by others.